\documentclass[fullpage,11pt]{article}

\usepackage{times}
\usepackage{courier}
\usepackage{amsmath}
\usepackage{graphicx}
\usepackage{color}
\usepackage{url}

\newcommand{\rao}[1]
           {\color{black} #1 \color{black}}

\newcommand{\nrao}[1]
           {\color{black} #1 \color{black}}

\newcommand{\nj}[1]
           {\color{black} #1 \color{black}}

\usepackage{clrscode3e}

\newcommand{\tfdu}{TFD$^{g_m}_{\textit{useful}}$}

\newcommand{\includefigure}[3][tbp]{
  \begin{figure}[#1]
    \centering
    \noindent
    \includegraphics{#2}
    \caption{#3}
    \label{figure:#2}
  \end{figure}
}

\def\mvp{}

\long\def\Ignore#1{}
\def\acronym#1{\text{\textsc\textbf{\small #1}}}

\newcommand{\aSet}[1]{{\left\{#1\right\}}}

\def\grad{\nabla}

\def\calG{\mathcal{G}}

\def\Begin{\Indentmore}
\def\Then{\kw{then} \Begin}

\def\CRIKEY3{\acronym{CRIKEY3}}

\def\eps{\varepsilon}

\newcommand{\citep}[1]{\citeauthor{#1}~\shortcite{#1}}

\def\Continue{\kw{continue}}

\title{Surrogate Search As a Way to Combat\\ Harmful Effects of
  Ill-behaved Evaluation Functions}

\author{William Cushing\thanks{An
    extended abstract of this paper appeared in the proceedings of SOCS
    2010. This research is supported in part by ONR grants
    N00014-09-1-0017 and N00014-07-1-1049, the NSF grant
    IIS-0905672, and by DARPA and the U.S. Army Research Laboratory
    under contract W911NF-11-C-0037.}
, J. Benton, Patrick Eyerich and Subbarao Kambhampati}

\begin{document}

\maketitle

\begin{abstract}
  \rao{ Recently, several researchers have found that cost-based
    satisficing search with A* often runs into problems. Although some
    ``work arounds'' have been proposed to ameliorate the problem,
    there has  been little concerted effort to pinpoint its origin. In
    this paper, we argue that the origins of this problem can be
    traced back to the fact that most planners that try to optimize
    cost also use cost-based evaluation functions (i.e., $f(n)$ is a
    cost estimate). We show that cost-based evaluation functions
    become ill-behaved whenever there is a wide variance in action
    costs; something that is all too common in planning domains.  The
    general solution to this malady is what we call a {\em surrogate
      search}, where a surrogate evaluation function that doesn't
    directly track the cost objective, and is resistant to
    cost-variance, is used. We will discuss some compelling choices for
    surrogate evaluation functions that are based on size rather than
    cost.  Of particular practical interest is a cost-sensitive
    version of size-based evaluation function — where the heuristic
    estimates the size of cheap paths, as it provides attractive
    quality vs. speed tradeoffs.}
\end{abstract}

\maketitle


\section{Introduction}
\label{intro}

Much of the scale-up and research focus in the automated planning
community in the recent years has been on satisficing planning.
Unfortunately, there has not been a concomitant increase in our
understanding of satisficing search.  Too often, the ``theory'' of
satisficing search in automated planning defaults to doing (W)A$^*$
with evaluation functions that optimize over the problem
objective. While this approach has shown itself to be empirically
better than some other techniques, there is a disconnect between the
behavior of the evaluation functions used in the search and the
computational efficiency we seek when performing satisficing
search. Finding a satisficing solution entails removing the
requirement of optimality in favor of better performance, but the
approach only goes half way; it removes the optimality requirement
without attention to efficiency. This turns out not to be too much of
a problem in domains with uniform cost values on each edge of the search
graph, but it can lead to a significant performance hit in problems
with non-uniform values. The issue is that the evaluation function can
be slow-rising with respect to the search depth, trapping the search
into performing an inordinate number of expansions. 

As we shall see,
these {\em ill-behaved evaluation functions} can cause arbitrarily
poor performance and crop up over a spectum of optimization problems
in automated planning; where we often want to consider non-uniform edge
costs in the search space. \nj{Problems that involve minimizing the
costs of actions in a plan (i.e., when performing cost-based
planning) offer a quintessential example of the issue, but many
other planning search spaces have this quality as well. Indeed, any
search space where the evaluation function fails to increase in
concordance with the search depth can suffer. As such, the common
objective in temporal planning of makespan minimization suffers an
equally dissatisfying performance drop when using makespan itself as the
evaluation function~\cite{benton10a}. This is because many search
decisions that involve adding actions in parallel to the current plan
will give zero increase in the evaluation
function. Because of this, most satisficing temporal planners already
more or less ignore the objective function during search, except to
prune plans with respect to incumbant solutions.}

To make our discussion more concrete, let us consider a cost-based
planning problem for which the cost-optimal and second-best solution
to a problem consist of 10 and 1000 unspecified actions respectively.
\emph{The optimal solution may be the longer one.}  Generally, a
combinatorial search can be seen to consist of two parts: a
``discovery'' part where the (optimal) solution is found and a
``proof'' part where the optimality of the solution is verified. An
A$^*$ search conflates these phases together and terminates only when
it picks the optimal path for expansion. So, how long should it take
just to discover the 10 action plan?  How long should it take to prove
(or disprove) its optimality?  In general (presuming PSPACE/EXPSPACE
$\ne$ P):
\begin{enumerate}
\item Discovery should require time exponential in, at most, 10.
\item Proof should require time exponential in, at least, 1000.
\end{enumerate}
That is, in principle, the only way to (domain-independently) prove
that the 10 action plan is better or worse than the 1000 action one is
to in fact go and discover the 1000 action plan.  Thus, A$^*$ search
with cost-based evaluation function will take time proportional to
$b^{1000}$ for either discovery or proof.  Simple breadth-first search
discovers a solution in time proportional to $b^{10}$ (and proof in
$O(b^{1000})$). 

We use both abstract and benchmark problems and show that this is a
systematic weakness of any search that uses an ill-behaved evaluation
function that grows arbitrarily slowly with respect to search
depth.
Using planning with action costs as the guiding example (without loss
of generality), we shall see that if $\eps$ is the smallest cost of an
action (after all costs are normalized so the most expensive action 
costs 1 unit), then the time taken to discover a depth $d$ optimal
solution will be $b^{\frac{d}{\eps}}$. If all actions have same cost,
then $\eps \approx 1$, whereas if the actions have significant cost
variance, then $\eps \ll 1$. For a variety of reasons, most real-world
planning domains do exhibit high cost variance, thus presenting an
``{\em $\eps$-cost trap}'' that forces any cost-based satisficing
search to dig its own ($\frac{1}{\eps}$ deep) grave.

We thus argue that the decision to use ill-behaved evaluation functions for
satisficing search should be reevaluated given their susceptibility to
$\eps$-cost traps, even when interested in that function's evaluation
of the resulting plan. But what exactly should take their place? We
suggest using a better-behaved {\em surrogate evaluation} function in
lieu of the ill-behaved one. There are two desiderata for such
surrogate evaluation:
\begin{enumerate}
\item they should be immune to the $\eps$-cost
traps by increasing at a reasonable rate with respect to search depth
and 
\item they should track the (true) objective well.
\end{enumerate}

In the case of
action costs, we will consider two size-based branch-and-bound
alternatives: the straightforward one which completely ignores costs
and sticks to a purely size-based evaluation function, and a more
subtle one that uses a cost-sensitive size-based evaluation function
(specifically, the heuristic estimates the size of the cheapest cost
path; see Section~\ref{setup}). We show that both of these outperform
cost-based evaluation functions in the presence of $\eps$-cost traps,
with the second one providing better quality plans (for the same run
time limits) than the first in our empirical studies. 

\nj{The issue takes a slightly different flavor when it comes to finding
satisficing solutions in temporal planning, where the usual stated
objective is makespan minimization. Balancing this with the avoidance
of $\eps$-cost traps takes great care, though temporal planners that
use heuristic search tend to limit use of the objective only for
pruning without directly using makespan for the heuristic value~(c.f.,
Temporal Fast Downward~\cite{eyerich09} and POPF~\cite{popf}).  We
seek to enhance such surrogate search approaches to pay more attention
to the makespan minimization objective. To these ends, we suggest the
use of a lookahead technique based on the usefulness of actions with
respect to the objective function.}


%






While some issues that arise with ill-behaved evaluation functions
have also been observed in the past (e.g.,
\cite{benton10a,lama-journal}), and some work-arounds have been
suggested, our main contribution is to bring to the fore their
fundamental nature; that the general phenomenon can be observed in any
planning problem where an evaluation function does not grow fast
enough with the search depth (e.g. measured in terms of node expansion
operations).

The rest of the paper is organized as follows. In the next section, we
present some preliminary notation, and formally introduce cost-based,
size-based as well as cost-sensitive size-based search alternatives
for searching with action costs.  Next, we present two abstract and
fundamental search spaces, which demonstrate that ill-behaved
evaluation functions are ``always'' needlessly prone to $\eps$-cost
traps (Section~\ref{epsilon-cost-trap}).  Section~\ref{topological}
strengthens the intuition behind this analysis by viewing best-first
search as flooding topological surfaces set up by evaluation
functions.  We will argue that of all possible topological surfaces
(i.e., evaluation functions) to choose for search, non-uniform ones
based on action costs are of the worst. \nj{In Section~\ref{surrogate-sec}
we propose a solution to this malady in terms of surrogate evaluation
functions. We describe two candidate surrogates for cost-based search
and a kindred approach for makespan minimization in temporal
planning.} In Section~\ref{sec:empirical}, we put all this analysis to empirical
validation by experimenting with \acronym{LAMA} \cite{lama-journal}
and SapaReplan \cite{sapa-replan} for searching with action costs, and Temporal Fast
Downward for searching for minimum makespan in a temporal setting.  The experiments do show
that surrogate search alternatives can out-perform previous,
ill-behaved evaluation functions.

\mvp
\section{Setup and Notation}
\label{setup}


We gear the problem set up to be in line with the prevalent view of
state-space search in modern, state-of-the-art satisficing planners. 
First, we assume the current popular approach of reducing planning to
graph search.  That is, planners typically model the state-space in a
causal direction, so the problem becomes one of extracting paths,
meaning whole plans do not need to be stored in each search node.
More important is that the structure of the graph is given {\em
  implicitly} by a procedure $\Gamma$, the child generator, with
$\Gamma(v)$ returning the local subgraph leaving $v$; i.e.,
$\Gamma(v)$ computes the subgraph $(N^+[v],E(\aSet{v},V-v)) = (\aSet{
  u \mid (v,u) \in E }+v, \aSet{ (v,u) \mid (v,u) \in E })$ along with all
associated labels, weights, and so forth. That is, our analysis
depends on the assumption that {\em an implicit representation of the
  graph is the only computationally feasible representation}, a common
requirement for analyzing the A$^*$ family of
algorithms~\cite{astar,dechter85}. 



The search problem is to find a path from an initial state, $i$, to
some goal state in $\calG$.  \nj{We use costs as a proxy for any single
evaluation objective (e.g., action costs or makespan).} and let them be
represented as edge weights, where $c(e)$ is the cost of the edge $e$.
Let $g_c^*(v)$ be the (optimal) cost-to-reach $v$ (from $i$), and
$h_c^*(v)$ be the (optimal) cost-to-go from $v$ (to the goal).  Then
$f^*_c(v) := g_c^*(v) + h_c^*(v)$, the cost-through $v$, is the cost
of the cheapest $i$-$\calG$ path passing through $v$.  For discussing
smallest solutions, let $f^*_s(v)$ denote the smallest $i$-$\calG$
path through $v$.  We can also consider the size of the cheapest
$i$-$\calG$ path passing through $v$, which we call
$\hat{f}^*_{s}(v)$.

We define a search node $n$ as equivalent to a path represented as a
linked list (of edges). In particular, we distinguish 
this from the state of $n$ 
(its last vertex), $\attribii{n}{v}$. We say $\attribii{n}{a}$ (for
action) is the last edge of the path and $\attrib{n}{p}$ (for parent)
is the subpath excluding $\attribii{n}{a}$.
\Ignore{Let $n_v$ denote the prefix of $n$ ending at $v$, and ${}_vn$ denote the
suffix starting at $v$.  Let $n''=n_vn'$ denote the path with prefix
$n_v$ and suffix ${}_vn'$.  Similarly l}Let $n'=na$ denote extending $n$
by the edge $a$ (so $a=(\attrib{n}{v},\attrib{n'}{v})$).  
The function $g_c(n)$ ($g$-cost) is just the recursive formulation of
path cost: $g_c(n) := 
g_c(\attribii{n}{p}) + c(\attribii{n}{a})$ ($g_c(n) := 0$ if $n$ is
the trivial path).  So 
$g_c^*(v) \le g_c(n)$ for all $i$-$v$ paths $n$, with equality for at
least one of them. Similarly let $g_s(n) := g_s(\attrib{n}{p}) + 1$
(initialized at 0), so that $g_s(n)$ is an upper bound on the shortest path
reaching the same state ($\attribii{n}{v}$).

A goal state is a target vertex where a plan may stop and be a valid
solution.  We fix a computed predicate $\calG(v)$ (a blackbox), the
\emph{goal}, encoding the set of goal states.  Let $h_c(v)$, the {\em
  heuristic}, be a procedure to estimate $h_c^*(v)$.  (Sometimes $h_c$
is considered a function of the search node, \emph{i.e.}, the whole
path, rather than just the last vertex.)  The heuristic $h_c$ is
called \emph{admissible} if it is a \emph{guaranteed} lower bound.  An
inadmissible heuristic lacks the guarantee, but might anyways be
coincidentally admissible.  Let $h_s(v)$ estimate the remaining depth
to the nearest goal, and let $\hat{h}_{s}(v)$ estimate the remaining
depth to the cheapest reachable goal.  Importantly, anything goes with
such heuristics.  Later we note that taking the size of a heuristic
that optimizes on cost is an
acceptable (and practical) interpretation of $\hat{h}_s(v)$.

We focus on two different definitions of $f$ (the evaluation
function). Since we study cost-based planning, we consider 

\begin{equation}
f_c(n) :=
g_c(n) + h_c(\attribii{n}{v})
\end{equation}
This is the (standard, cost-based)
\nj{{\em ill-behaved evaluation function}} of $A^*$:
cheapest-completion-first.  We compare this to 
\begin{equation}
f_s(n) := g_s(n) +
h_s(\attribii{n}{v})
\end{equation}
This is the canonical size-based (or search distance)
{\nj {\em surrogate evaluation function}}, equivalent to $f_c$ under uniform
weights. Any combination of $g_c$ and $h_c$ is {\em cost-based}; any
combination of $g_s$ and $h_s$ is {\em size-based} (\emph{e.g.},
breadth-first search is size-based).  The evaluation function
$\hat{f}_s(n) := g_s(n) + \hat{h}_s(\attribii{n}{v})$ is also
size-based, but cost-sensitive.

Pseudo-code for best-first branch-and-bound search of implicit
graphs is shown below.  
It continues searching after a solution is encountered and uses the
current best solution value to prune the search space
(line~\ref{line:branch-and-bound}).  The search is performed on a
graph implicitly represented by $\Gamma$, with the assumption being
that the explicit graph is so large that it is better to invoke
expensive heuristics (inside of \proc{evaluate}) during the search than it is to
just compute the graph up front. 
The question considered by this paper is how to implement
\proc{evaluate}.

{\footnotesize
\begin{codebox}
 \Procname{$\proc{Best-First-Search}(i,\calG,\Gamma,h_c)$}
\li  initialize search
\li  \While \id{open} not empty \Do
\li      $n \gets \attribii{open}{\func{remove}}()$
\li      \If $\proc{bound-test}(n,h_c)$ \Then \Continue \End \label{line:branch-and-bound}
\li      \If $\proc{goal-test}(n,\calG)$ \Then \Continue \End 
\li      \If $\proc{duplicate-test}(n)$ \Then \Continue \End 
\li      $s \gets \attribii{n}{v}$
\li      $\id{star} \gets \Gamma(s)$ \RComment{Expand $s$}
\li      \For each edge $a=(s,s')$ from $s$ to a child $s'$ in $\id{star}$ \Do
\li          $n' \gets n a$
\li          \fbox{$f \gets \proc{evaluate}(n')$}\rule[-8pt]{0pt}{20pt}
\li          $\attribii{open}{\func{add}}(n',f)$
         \End
     \End
\li  \Return $\id{best-known-plan}$ \RComment{Optimality is proven.}
\end{codebox}
}


With respect to normalizing costs, we can let $\eps:=\frac{\min_a
  c(a)}{\max_a c(a)}$, that is, $\eps$ is the least cost edge after
normalizing costs by the maximum cost (to bring costs into the range
$[0,1]$). 
We use the symbol $\eps$ for this ratio as we anticipate actions with
high cost variance in real world planning problems.  For example:
boarding versus flying (ZenoTravel), mode-switching versus machine
operation (Job-Shop), and (unskilled) labor versus (precious) material
cost. 




\begin{figure}[h]
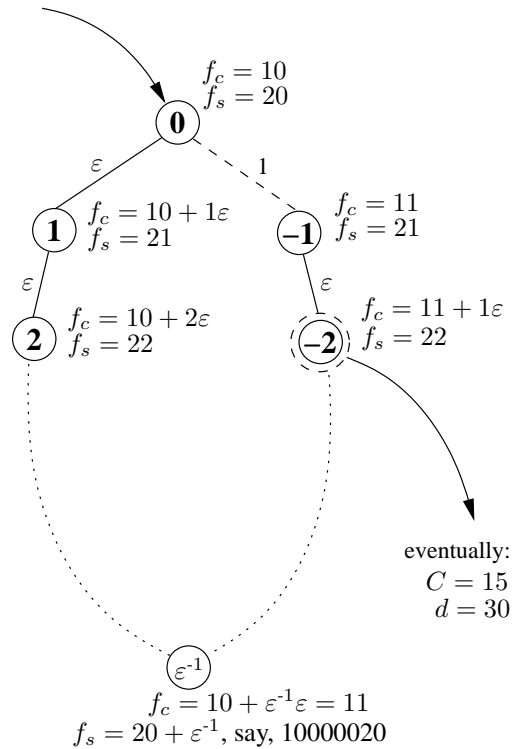

\centering
\include{trap}
\vspace{-0.15in}
\caption{\label{fig:cycle-trap}An $\eps$-cost trap for cost-based
  search. 
}
\vspace{-0.15in}
\end{figure}

\mvp
\section{ Two Canonical Case of $\eps$-cost Trap}
\label{epsilon-cost-trap} 

In this section we argue that the mere presence of $\eps$-cost edge
weights misleads search, and that this is not an accidental
phenomenon, but a systemic weakness of the very concept of
``\nj{(ill-behaved)} cost-based evaluation functions + systematic search +
combinatorial graphs''.
We base this analysis in two abstract search spaces, in order to
demonstrate the fundamental nature of such traps. 


\begin{figure*}
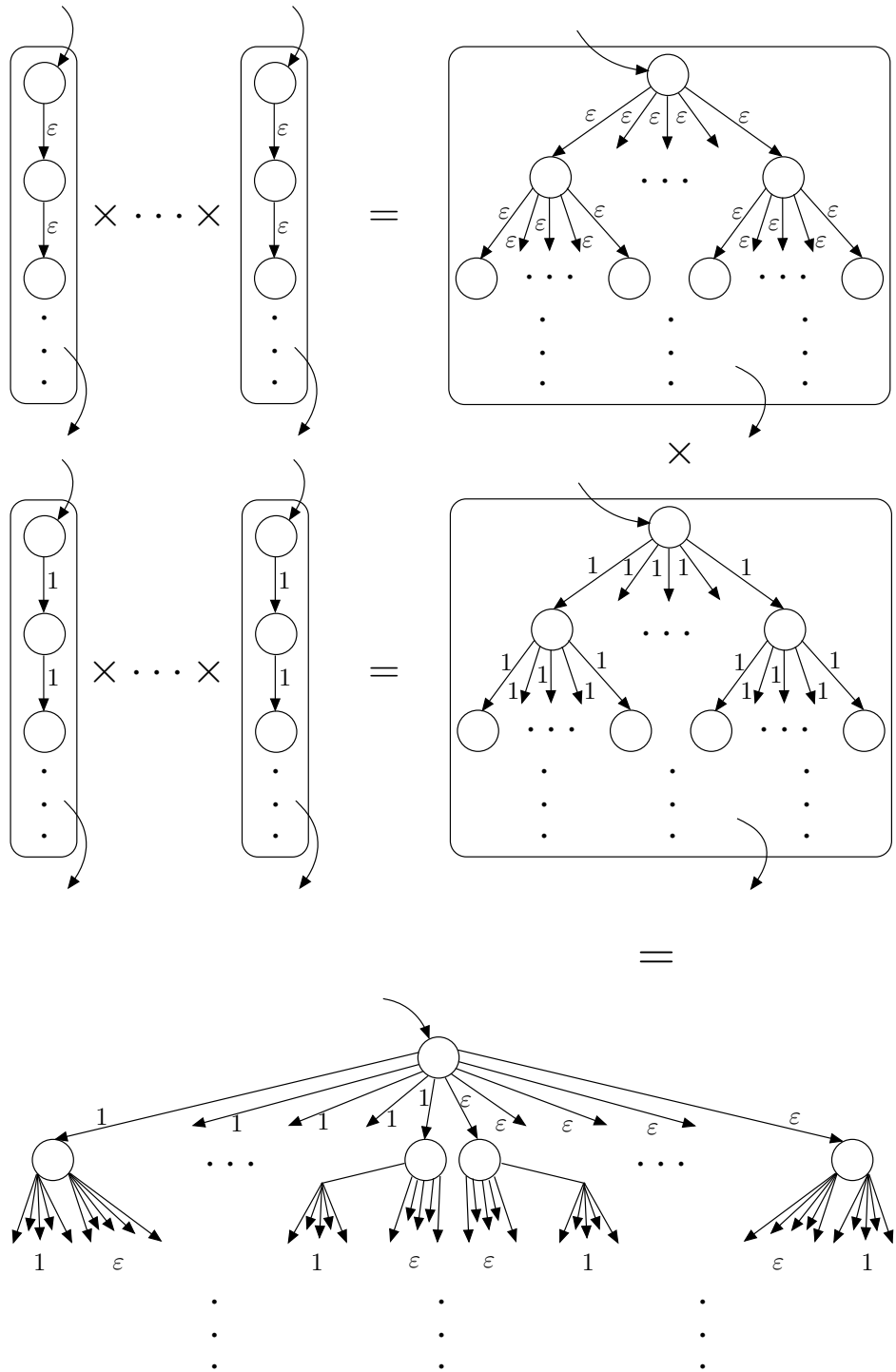

\centering
\include{tree}
\vspace{-0.1in}
\caption{\label{fig:tree-trap}A trap for cost-based search involving
  exponentially large subtrees on $\eps$-cost edges. 
}
\vspace{-0.15in}
\end{figure*}

For our analysis, the first abstract space we consider is a simple,
non-trivial, non-uniform cost, intractably large, search space: the
search space of an enormous cycle with one expensive edge.  The second
abstract space we consider is a more natural model of search (in
planning): a uniform branching tree.  Traps in these spaces are 
exponentially sized \emph{and connected} sets of $\eps$-cost edges:
not the common result of, for example, a typical random model of
search. We briefly consider why planning benchmarks naturally give
rise to such structure.

\mvp
\subsection{Cycle Trap}

In this section we consider the simplest abstract example of the
$\eps$-cost `trap'.  The notion is that applying increasingly powerful
heuristics, domain analysis, learning techniques, \dots, to one's
search problem transforms it into a simpler `effective graph' --- the
graph for which Dijkstra's algorithm~\cite{dijkstra-search} produces
isomorphic behavior.  For example, let $c'$ be a new edge-cost
function obtained by setting edge costs to the difference in $f$
values of the edge's endpoints: Dijkstra's algorithm on $c'$ is A$^*$
on $f$.\footnote{Systematic inconsistency of a heuristic translates to
  analyzing the behavior of Dijkstra's algorithm with many
  \emph{negative} `cost' edges, a typical reason to assume consistency
  in analysis.}  Similarly take $\Gamma'$ to be the result of applying
one's favorite incompleteness-inducing pruning rules to $\Gamma$ (the
child generator), say, helpful actions~\cite{ff}; then Dijkstra's
algorithm on $\Gamma'$ is A$^*$ with helpful action pruning.

We presume the effective search graph remains very complex despite
clever inference (or there is nothing to discuss). If there is a
problem with search behavior in an exceedingly simple graph then we
can suppose that no amount of domain analysis, learning, heuristics,
and so forth, will incidentally address the problem: such inference
must specifically address the issue of non-uniform costs.  When none
of the bells and whistles consider non-uniform costs to be a serious
issue, the search permits wildly varying edge ``costs'' even in the
effective search graph: $\eps\approx \eps'=\frac{\min_e c'(e)}{\max_e
  c'(e)}$.  We demonstrate that this by itself is enough to produce
very troubling search behavior: $\eps$-cost is a fundamental challenge
to be overcome in planning.

There are several candidates for simple non-trivial state-spaces
(\emph{e.g.}, cliques), but clearly the cycle is fundamental (what
kind of `state-space' is acyclic?).  So, the state-space we consider
is the cycle, with associated exceedingly simple metric consisting of
all uniform weights but for a single expensive edge.
Its search space is certainly a simple non-trivial search space: the rooted
tree on two leaves.  So the single unforced decision to be made is in which
direction to traverse the cycle: clockwise or counter-clockwise. 

\noindent{\bf $\varepsilon$-cost Trap:} Consider a counting problem of
making some variable, $x$, encoded in $k$ bits represent
$2^{k}-2\equiv -2 \pmod{2^k}$, starting from $0$, using only the
operations of increment and decrement. We illustrate the search in
Figure~\ref{fig:cycle-trap}. There are 2 minimal solutions:
incrementing $2^k-2$ times, or decrementing twice.  Set the cost of
incrementing and decrementing to 1, except for transitioning between
$x\equiv 0$ and $x\equiv-1$ costs, say, $2^{k-1}$ (in either
direction).  Then the 2 minimal solutions cost $2^k-2$ and
$2^{k-1}+1$, or, normalized, $2(1-\varepsilon)$ and $1+\varepsilon$.
Cost-based search loses: While both approaches prove optimality in
exponential time ($O(2^k)$), size-based search discovered that optimal
plan in constant time.  

The heuristic perceives all movement on the cycle to be irrelevant to
achieving high quality plans.  The state with label \textbf{-2} is one
interesting way to leave the cycle, there may be (many) others.  $C$
denotes the cost of one such continuation from \textbf{-2}, and $d$
its depth.  Edge weights nominally denote changes in $f_c$: as given,
locally, these are the same as changes in $g_c$.  But increasing $f_s$
by 1 at \textbf{-1} (and descendants) would, for example, model
instead the special edge as having cost $\frac{1}{2}$ and being
perceived as worst-possible in an undirected graph.

\def\paragraph#1{\noindent{\bf #1}}

\smallskip\paragraph{Performance Comparison: All Goals.}  The goal
$x\equiv -2$ is chosen to best illustrate the trap. Consider the
discovery problem for other goals.  With the goal in the interval $2^k
\cdot [0,\frac{1}{2}]$ cost-based search is twice as fast.  With the
goal in the interval $2^k \cdot [\frac{1}{2},\frac{2}{3}]$ the
performance gap narrows to break-even.  For the last interval,
$2^k\cdot[\frac{2}{3},1\rangle$, the size-based approach takes the
lead --- \emph{by an enormous margin}.  There is one additional region
of interest.  Taking the goal in the interval
$2^k\cdot[\frac{2}{3},\frac{3}{4}]$ there is a trade-off: size-based
search finds a solution before cost-based search, but cost-based
search finds the optimal solution first.  Concerning time till
optimality is proven, the cost-based approach is monotonically faster
(of course).  Specifically, the cost-based approach is faster by a
factor of 2 for goals in the region $2^k\cdot [0,\frac{1}{2}]$, not
faster for goals in the region $2^k\cdot[\frac{3}{4},1\rangle$, and by
a factor of $(\frac{1}{2}+2\alpha)^{-1}$ (bounded by 1 and 2) for
goals of the form $x\equiv 2^k(\frac{1}{2}+\alpha)$, with $0 < \alpha
< \frac{1}{4}$.

\smallskip\paragraph{Performance Comparison: Feasible Goals.}
Considering all goals is inappropriate in the satisficing context; to
illustrate, consider $k=1000$ as an example of a large value of $k$.
Fractions of exponentials are still exponentials --- even the most
patient reader working out this example will have forcibly terminated
either search \emph{long} before receiving any useful output.  Except
if the goal is of the form $x\equiv 0 \pm f(k)$ for some
sub-exponential $f(k)$.  Both approaches discover (and prove) the
optimal solution in the positive case in time $O(f(k))$ (with
size-based performing twice as much work).  In the negative case, only
the size-based approach manages to discover a solution (the optimal
one, in time $O(f(k))$) before being killed.  Moreover, while it will
fail to produce a proof of such before termination, based on our
understanding of the domain, we can show it to be posthumously
correct.  ($2^k-f(k) > 2^k\cdot\frac{3}{4}$ for any sub-exponential
$f(k)$ with large enough $k$.)

\smallskip\paragraph{How Good is Almost Perfect Search Control?} Keep
in mind that the representation of the space as a simple $k$ bit
counter is \emph{not} available.  In particular what `increment'
actually stands for is an inference-motivated choice of a single
operator out of a large number of executable and promising operators
at each state --- in the language of Markov Decision Processes, we are
allowing inference to be so close to perfect that the optimal policy
is known at all but 1 state.  Only one decision remains \dots but no
methods cleverer than search remain.  Still the graph is intractably
large.  Cost-based search only explores in one direction: left, say.
In the satisficing context such behavior is entirely inappropriate.
What is appropriate?  We want the search to explore the space so that
a solution that exists only one step to the right can still be found,
even if it is not optimal.


\Ignore{There is only one decision left to be
made \dots\ but we have run out of any methods more clever than search.
Still the effective search graph is intractably large.  What should be
done?  A great research problem.  What shouldn't be done?  Pretend
that cost is a non-issue, specifically, asserting that $A^*$ is
complete so long as costs are strictly positive is not a particularly
helpful observation --- $\eps$ is only negligibly better than 0.
Especially writing pseudocode for it by manipulating edge weights
rather than '1' is not helpful.  Keep in mind that one \emph{says}
$A^*$/best-first-search and applies super-expensive heuristics (rather than writing
along the lines of ``transform the graph $G$ to $G'$ by the
manipulation $g'=g+h$'') in order to emphasize the analysis framework introduced to
talk about $A^*$: the graph is not accessible in full, ever.}

\mvp
\subsection{Branching Trap}

In the counter problem the trap is not even \emph{combinatorial}; the
search problem consists of a single decision at the root, and the trap
is just an exponentially deep path.  For example, appending a large
enough Towers of Hanoi problem to a planning benchmark, setting its
actions at $\eps$-cost, will hurt cost-based search --- even given the
perfect heuristic for the puzzle! Besides Hanoi, though,
exponentially deep paths are not typical of planning benchmarks.  So
in this section we demonstrate that exponentially large subtrees on
$\eps$-cost edges are also traps.

Consider $x>1$ high cost actions and $y>1$ low cost actions in a
uniform branching tree model of search space.  The model is
appropriate up to the point where duplicate state checking becomes
significant.  See the example illustrated in
Figure~\ref{fig:tree-trap}. 

We have two rather distinct kinds of physical objects that exist in
the domain with primitive operators at rather distinct orders of
magnitude; supposing uniformity and normalizing, one type involves
$\eps$-cost and the other involves cost $1$.  In this case, there is a
low-cost subspace, a high-cost subspace, and the full space, where
each is a uniform branching tree.  As trees are acyclic, it is
probably best to think of these as search (rather than state) spaces.
As depicted, planning for an individual object is trivial as there is
no choice besides going forward. Other than that no significant amount
of inference is being assumed, and in particular the effects of a
heuristic are not depicted.  For cost-based search to avoid ``dying''
due to memory or time computational resources, the heuristic would
need to forecast every necessary cost 1 edge, so as to reduce its
weight closer to 0.  (Note that the aim of a heuristic is to drive all
the weights to 0 along optimal/good paths, and to infinity for
not-good/terrible/dead-end choices.)  If any cut of the space across
such edges (separating good solutions) is not foreseen, then
backtracking into {\em all} of the low-cost subspaces so far
encountered commences, to multiples of depth $\eps^{-1}$ --- one such
multiple for every unforeseen cost 1 cut.

Observe that in the object-specific
subspaces (the paths), a single edge ends up being multiplied into
such a cut of the global space. Suppose the solution of interest costs
$C$, in normalized units, so the solution lies at depth $C$ or
greater.  Then cost-based search faces a grave situation:
$O((x+y^{\frac{1}{\eps}})^C)$ possibilities will be explored before
considering all potential solutions of cost $C$.

A size-based search only ever considers at most $O((x+y)^d)=O(b^d)$
possibilities before consideration of all potential solutions of size
$d$.  But the more interesting question is how long it takes to find
solutions of fixed cost rather than fixed depth.  Note that
$\frac{C}{\eps} \ge d \ge C$.  Assuming the high cost actions are
relevant, that is, some number of them are needed by solutions, we
have that solutions are not actually hidden as deep as
$\frac{C}{\eps}$.  To help see this, suppose that solutions tend to be
a mix of high and low cost actions in equal proportion.  Then the
depth of those solutions with cost $C$ is $d=2\,\frac{C}{1+\eps}$
(\emph{i.e.}, $\frac{d}{2} \cdot 1 + \frac{d}{2} \cdot \eps = C$).  At
such depths the size-based approach is the clear winner:
$O((x+y)^{\frac{2C}{1+\eps}}) \ll O((x+y^{\frac{1}{\eps}})^C)$
(normally).

Consider the case where $x=y=\frac{b}{2}$, then:
\begin{align*}
\text{size effort} / \text{cost effort} &\approx\\
b^{\frac{2C}{1+\eps}} / \left(x + y^{\frac{1}{\eps}}\right)^C 
&< b^{\frac{2C}{1+\eps}} / y^{\frac{C}{\eps}},\\
&< 2^{\frac{C}{\eps}} / b^{\frac{C}{\eps}\frac{1-\eps}{1+\eps}},\\
&< \frac{2}{b^{\frac{1-\eps}{1+\eps}}}^{\frac{C}{\eps}},
\end{align*}
and, provided $\eps < \frac{1-\log_b 2}{1+\log_b 2}$ (for $b=4$,
$\eps < \frac{1}{3}$), the last is always less than 1 and, for that
matter, goes quickly to 0 as $C$ increases and/or $b$ increases and/or $\eps$ decreases.

Generalizing from this example, the size-based approach is faster at
finding solutions of any given cost, as long as (1) high-cost actions
constitute at least some constant fraction of the solutions considered
(high-cost actions are relevant), (2) the ratio between high-cost and
low-cost is sufficiently large, (3) the effective search graph (post
inference) is reasonably well modeled by an infinite uniform branching
tree (i.e., huge enough to render duplicate checking
negligible, or at least not especially favorable to cost-based
search), and most importantly, (4) the cost function in the effective
search graph \emph{still} demonstrates a sufficiently large ratio
between high-cost and low-cost edges (no inference has attempted to
compensate).




%
%
%
\mvp
\section{Search Topology and Satisficing Solutions}
\label{topological}

We view evaluation functions ($f$) as topological surfaces over search
nodes, so that generated nodes are visited in, roughly, the order of
$f$-altitude. With non-monotone evaluation functions, the set of
nodes visited before a given node is all those contained within some
basin of the appropriate depth --- picture water flowing from the
initial state: if there are dams then such a flood could temporarily
visit high altitude nodes before low altitude nodes.  (With very
inconsistent heuristics --- large heuristic weights --- the metaphor
loses explanatory power, as there is nowhere to go but downhill.)

\Ignore{If we take a single point inside such a basin (but not one defining
the brim) and alter its altitude over the entire range of that basin's
depth, we will not have changed the set of nodes inundated prior to
the brim.  If there were no solutions prior to the brim, then we will
not have altered any externally visible behavior of the search:
Whenever best-first search finally finds a solution it will no longer
have mattered how all the prior nodes were ordered.  To illustrate,
$\text{IDA}^*$ deserves its name, despite exploring the space in an entirely
different order from A$^*$ in its last iteration.  Similarly one can
simulate any best-first search by choosing to explore in any other
order, pruning anything worse than a conjectured bound (and iterating
the conjectures).
}

All reasonable choices of search topology will eventually lead to
identifying and proving the optimal solution (e.g., assume
finiteness, or divergence of $f$ along infinite paths).  Some search
toplogies will
produce a whole slew of suboptimal solutions along the way, eventually
reaching a point where one begins to wonder if the most recently
reported solution is optimal.  Others report nothing until finishing.
The former are \emph{interruptible}~\cite{zilberstein98a}, which is
one way to more formally define satisficing.\footnote{Another way that
  Zilberstein suggests is to specify a \emph{contract}; the 2008 and
  2011 planning competitions have such a format~\cite{ipc08}.}
Admissible cost-based topology is the least interruptible choice: the
only reported solution is also the last path considered.  Define the
cost-optimal footprint as the set of plans considered.  Gaining
interruptibility is a matter of raising the altitude of large portions
of the cost-optimal footprint in exchange for lowering the altitude of
a smaller set of non-footprint search nodes --- allowing sub-optimal
solutions to be considered.  
Note that interruptibility comes at the
expense of total work.

So, along the lines of confirming the intuition that interruptibility
is a reasonable notion of satisficing: the cost-optimal approach is a
poor choice for satisficing solutions.  Said another way, proving
optimality is about increasing the lower bound on true value, while
solution discovery is about decreasing the upper bound on true value.
It seems appropriate to assume that the fastest way to decrease the
upper bound is more or less the opposite of the fastest way to
increase the lower bound --- with the notable exception of the very
last computation one will ever do for the problem: making the two
bounds meet (proving optimality).

Intuitively the search should not be completely blind to costs: just
defensive about possible $\eps$-traps. For size-based topology, with
respect to any cost-based variant, the `large' set is the set of
\emph{longer} yet \emph{cheaper} plans, while the `small' set is the
\emph{shorter} yet \emph{costlier} plans.  In general one expects
there to be many more longer plans than shorter plans in combinatorial
problems, so that the increase in total work is small, relative to the
work that had to be done eventually (exhaust the many long, cheap,
plans).  The additional work is considering exactly plans that are
costlier than necessary (potentially suboptimal solutions).  The idea
of the trade-off is good, but even the best version of a purely
size-based topology will not be the best trade-off possible.

Not finding the cheapest path first comes with the price of
re-expansion, so the satisficing intent comes hand in hand with
re-expansion of states. 
Indeed, duplicate detection and re-expansion are, in practice,
important issues. Besides the obvious kind of re-expansion that
$\text{IDA}^*$~\cite{korf85a} performs between iterations, it is also
true that it considers paths which A$^*$ never would (even subsequent
to arming $\text{IDA}^*$ with a transposition table) --- it is not
really true that one can reorder consideration of paths however one
pleases.  In particular at least some kind of breadth-first bias is
appropriate, so as to avoid finding woefully suboptimal plans to
states early on, triggering giant cascades of re-expansion later on.

\section{Handling $\eps$-Cost Traps with Surrogate Search}
\label{surrogate-sec}

\nj{
We propose that a principled way to combat
  the effects of $\eps$-cost traps is to swap the objective-sensitive 
evaluation function, which is  ill-behaved, with a surrogate
evaluation function that is well behaved, but does not directly track
the objective. While surrogate functions avoid the $\eps$-cost traps,
they do so at the expense of solution quality. To make up for this, we will consider
branch-and-bound search regimes, where the search can continuously
improve the quality of the solution. 
Obviously, for a given objective, there can be multiple
surrogate evaluation functions that differ in terms of how
closely/loosely they track the obejctive. They offer a spectrum of 
computation vs. quality tradeoffs--with the functions that track the
objective function more closely converging on higher quality
solutions, but taking considerably longer time to find even one
solution. The art here is to strike a good balance in this tradeoff.
Specifically, the challenge is in finding effective surrogate
  evaluations that increase in accordance with search depth while
  navigating the tradeoff between defending against $\eps$-cost traps
  and focusing search on the objective. 

%

To concretely illustrate the art of picking good surrogate evaluation
functions, we consider two distinct planning problems--a cost-based
planning planning problem, as well as a temporal planning problem
involving makespan minimization. In the former case, we first show
that a purely sized-based surrogate evaluation function is able to
avoid the search cost imposed by $\eps$-cost trap. We then describe a
more sophisticated surrogate evaluation function that estimates the
size of a cost-optimal plan, and show that it has better performance
in terms of solution quality. The empirical support for these claims
is provided by experiments run on LAMA, a state of the art planner.
For the makespan optimizing temporal planning, most practical methods,
such as Temporal Fast Downward, already abandon objective functions
directly tracking makespan. By viewing this decision in terms of the
surrogacy of the evaluation function, we propose a method for further
improving cost-vs-quality tradeoff that involves performing a
lookahead based on the usefulness of actions according to the
objective function.


}


\subsection{Surrogate Evaluation Functions for Cost-based Planning}
For cost-based planning we consider two kinds of surrogate evaluation
functions: the first is purely based on size, while the second is also
sensitive to costs. 

\medskip\medskip
\paragraph{Pure Size-Based Evaluation Function:} Here we replace
cost-based evaluation function with a pure size-based one. That is, we
search with $f$ value being the size (number of actions) in the
solution, even though we are interested in optimizing the cost of the
solution. In particular, the heuristic will estimate the size of the
shortest length path from the current state.  By using size-based
evaluation function, we effectively force $\eps$ to be 1. Since the
search is not cost-focused, the first solution it finds may not
necessarily have high quality, however the search will avoid
$\eps$-cost traps. We handle quality of solutions  by employing a branch-and-bound
search.

\medskip\medskip
\paragraph{Cost-Sensitive Size-Based Evaluation Function:} This is
an improvement of the pure-size based evaluation function that
improves its quality focus while still retaining $\eps$ being
1. Specifically, the evaluation function here is still measured in
terms of size or number of actions, but the heuristic aims to estimate
the size of the cheapest {\em cost} path from the current state. By
tracking the cheapest cost path, the evaluation function becomes more
quality focused, and by measuring that path in terms of the number of
actions (rather than the cumulative cost), it avoids the $\eps$-cost
trap. Once again, the quality of the solution can be improved with a
branch-and-bound search regime. To elaborate further, the
cost-sensitive size-based evaluation functions can use the standard
relaxed-plan heuristics that select the relaxed plan itself in terms
of the cost \cite{minh-cost-plan}. However, rather than take the 
cost of the relaxed plan, we take the {\em size} of the relaxed plan,
computing $\hat{h}_s$, and use $\hat{f}_s = g_s + \hat{h}_s$ for the
evaluation function (see Section~\ref{setup}).

\subsection{Surrogate Evaluation Functions for Temporal Planning}

\nj{ In the case of temporal planning, we look toward enhancing search
  methodologies that already do not directly search over the objective
  of makespan minimization. To do this, we perform a lookahead over
  {\em useful search opeators}. We define useful search operators as
  those whose absence would lead to a worse quality solution.  They
  relate to the idea of useless operators defined by Wehrle, et
  al.~\cite{wehrle-etal:icaps2008}; indeed, we can find a useful
  operator by using the same method for finding useless operators. We
  say an operator $o$ is {\em useful} in a state $s$ if
  $d^{\bar{o}}(s) > d(o(s))$, where $d^{\bar{o}}(s)$ is the optimal
  distance to a goal state without an operator $o$ (i.e., $d(s)$ and
  $d^{\bar{o}}(s)$ are perfect heuristics).  We say the state $o(s) =
  s'$ is useful if the operator generating it is useful and we can use
  this notion to enhance satisficing planners that use best-first
  search. We define the {\em heuristic degree of usefulness} of an
  operator $o$ as $\upsilon^{o}(s) = h^{\bar{o}}_m(s) - h_m(o(s))$.
  With this information, we create a new search approach that
  interleaves local search decisions using degree of usefulness on
  makespan when we detect a $g$-value plateau (i.e., no change in
  makespan). In other words, we supplement the surrogate evaluation
  function with a one-step lookahead procedure. This procedure
  generates a set of child states for the best node on the queue and
  calculates their ``makespan-usefulness'' using a heuristic on
  makespan.  The procedure  expands the most
  makespan-useful node (unless no nodes are makespan-useful), then
  puts all expanded nodes back into the best-first search queue,
  ordered by the surrogate function. }

\mvp

\section{Empirical Evaluation}
\label{sec:empirical}

\nj{In this section, we verify that  the effect of
  $\eps$-cost traps in ill-behavioed evaluation functions is
  disruptive to planner performance, both in cost-based planning and
  temporal planning. We shall see that the use of size-based surrogate
  search does better in the case of cost-based planning.  We will also
  see the effect in temporal planning in the planning framework of
  Temporal Fast Downward, discuss Temporal Fast Downward's surrogate
  search and see how our lookahead technique improves plan quality
  over it.}

\begin{figure}[t]
\begin{center}
\includegraphics[clip=true,trim=0in 0.6in 0in 0in,width=2.7in,height=1.8in]{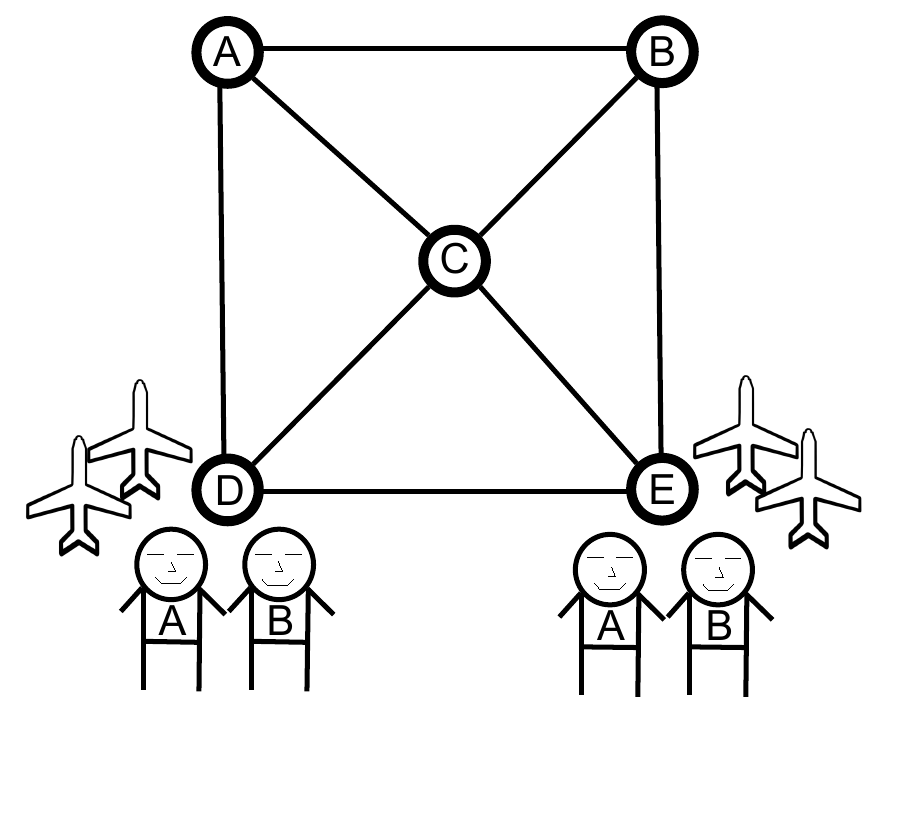}
\end{center}
\caption{\label{fig:travel}Rendezvous problems. Diagonal edges cost
  7,000, exterior edges cost 10,000. Board/Debark cost 1.\vspace{-0.15in}}
\end{figure}

We demonstrate existence of the problematic planner behavior when
using ill-behaved evaluation functions in both \acronym{LAMA-2008} for
cost-based planning and Temporal Fast Downward for temporal
planning.
 In \acronym{LAMA-2008} we use problems in the travel domain
(simplified ZenoTravel, zoom and fuel removed), as well as two other
IPC domains. In Temporal Fast Downward we use domains from the
International Planning Competition of 2008 (IPC-2008).  Analysis of
\acronym{LAMA} is complicated by many factors, so we also test the
behavior of SapaReplan on simpler instances (but in all of
ZenoTravel).  The first set of problems concern a rendezvous at the
center city in the location graph depicted in Figure~\ref{fig:travel};
the optimal plan arranges a rendezvous at the center city.  The second
set of problems is to swap the positions of passengers located at the
endpoints of a chain of cities.

\mvp

\subsection{Cost-Based Planning}
\subsubsection{Evaluation of Cost-based planning on LAMA}

In this section we demonstrate the performance problem wrought by
$\eps$-cost in a state-of-the-art (2008) planner ---
\acronym{LAMA}~\cite{lama-journal}, the leader of the cost-sensitive
(satisficing) track of IPC'08~\cite{ipc08}.  With a completely trivial
recompilation (set a flag) one can make it ignore the given cost
function, effectively searching by $f_s$, i.e., pure-size. This
methodology was evaluated
by Richter and Westphal~(\cite{lama-journal}). With slightly
more work one can do better and have it use $\hat{f}_s$ as its
evaluation function, i.e., use a cost-sensitive heuristic estimate of
$\hat{d}$ and allow the search be size-based, but still compute costs
correctly for branch-and-bound. We call this latter modification
\acronym{LAMA}-size.  Ultimately, the observation is that
\acronym{LAMA}-size outperforms \acronym{LAMA} --- an astonishing feat
for such a trivial change in implementation.

\acronym{LAMA}\footnote{Options: `fFlLi'.} defies analysis in a number
of ways: \emph{landmarks},\Ignore{\footnote{``Landmark analysis'' is a
    misnomer: in standard usage a landmark is a \emph{sufficient}
    condition (for either dead-ends, ``\dots then you've gone too
    far'', or achievement, ``\dots then you can't miss it''), but in
    automated planning the jargon stands only for a \emph{necessary}
    condition.  Principled use of these, as \acronym{LAMA} does, is
    then not an \emph{especially} promising attack upon search
    plateaus.}} \emph{preferred operators}, \emph{dynamic evaluation
  functions}, \emph{multiple open lists}, and \emph{delayed
  evaluation}, all of which effect potential search plateaus in
complex ways.  Nonetheless, it is essentially a cost-based approach.

\Ignore{
For the first several stages the
search space is that induced by only \emph{preferred operators},
afterwords, all operators (actions) are considered.  Eventually the heuristic
weight decreases to 1, and remains constant in all later stages; there
are potentially multiple such stages as the heuristic(s) are not
(assumed to be) admissible.\footnote{Restarting is one approach to
  branch-and-bound; restarting is particularly appropriate if the evaluation
  function changes upon finding solutions.   Once the heuristic weight
  is constant, though, it may no longer make sense to restart.}
\acronym{LAMA} uses \emph{multiple open lists} (two, with differing heuristics), with search nodes being
entered into every open list, avoiding needless re-expansion by exploiting the closed list.  (So duplicate
states may induce re-expansion, if the cheaper path is found second,
but not duplicate search nodes.)  \acronym{LAMA} also employs \emph{delayed evaluation}, meaning
that the $f$-value actually assigned is based on the $g$-value of the
node in question and the $h$-value of its parent (so the search
assumes no child makes progress w.r.t.\ $h$).  As a result, siblings are visited
cheapest first.}

\noindent{\bf Results.}\footnote{New best plans
  for Elevators were found (largely by \acronym{LAMA}-size).  The baseline planner's score is 71.8\% against the better reference plans.}  With more than about 8 total passengers, \acronym{LAMA}
is unable to complete any search stage except the first (the greedy search). 
For the same problems, \acronym{LAMA}-size finds the same first plan (the
heuristic values differ, but not the structure), but is then subsequently able to
complete further stages of search.  In so doing it sees marked
improvement in cost; on the larger problems this is due only to
finding better variants on the greedy plan.  Other domains are
included for broader perspective, woodworking in particular was chosen
as a likely counter-example, as all the actions concern just one
type of physical object and the costs are not wildly different.  For
the same reasons we would expect \acronym{LAMA} to out-perform \acronym{LAMA}-size in some
cost-enhanced version of Blocksworld.

\begin{table}
  \caption{Percentage IPC score improvement on \acronym{LAMA} variants.\label{tbl:lama}}{%
\begin{tabular}{|c||c|c|}
\hline
Domain & \acronym{LAMA} & \acronym{LAMA}-size\\
\hline
Rendezvous              & 70.8\% &83.0\%\\
Elevators               & 79.2\% &93.6\%\\
Woodworking             & 76.6\% &64.1\%\\
\hline
\end{tabular}
}
\end{table}

\mvp
\subsubsection{Evaluation of Cost-Based Planning on SapaReplan}

We also consider the behavior of SapaReplan on the simpler set of
problems.
This planner is much less sophisticated in terms of
its search than \acronym{LAMA}, in the sense that it does not use dual
queues or lazy evaluation.  The problem is just to swap the locations
of passengers located on either side of a chain of cities.  A plane
starts on each side, but there is no actual advantage to using more
than one (for optimizing either of size or cost): the second plane
exists to confuse the planner.  Observe that smallest and cheapest
plans are the same.  So in some sense the concepts have become only
superficially different; but this is just what makes the problem
interesting, as despite this similarity, still the behavior of search
is strongly affected by the nature of the evaluation function.  We
test the performance of $\hat{f}_s$ and $f_c$, as well as a hybrid
evaluation function similar to $\hat{f}_s + f_c$ (with costs
normalized).  We also test hybridizing via tie-breaking conditions,
which ought to have little effect given the rest of the search
framework.

\noindent{\bf Results.}\footnote{The results differ markedly between the 2 and 3 city sets of problems
because the sub-optimal relaxed plan extraction in the 2-cities problems
coincidentally produces an essentially perfect heuristic in many of
them.  One should infer that the solutions found in the 2-cities
problems are sharply bimodal in quality and that the meaning of the average is
then significantly different than in the 3-cities problems.} The
size-based evaluation functions find better cost plans faster (within
the deadline) than cost-based evaluation functions.  The hybrid
evaluation function also does relatively well, but not as well as
could be hoped. Tie-breaking has little effect, sometimes negative.

We note that Richter and Westphal (2010) also report
that replacing cost-based evaluation function with a pure size-based one
improves performance over \acronym{LAMA} in multiple other domains. Our version
of \acronym{LAMA}-size uses a cost-sensitive size-based search
($\hat{h}_s$), and our results,
in the domains we investigated, seem to show bigger improvements over
the size-based variation on \acronym{LAMA} obtained by
\emph{completely} ignoring costs ($h_s$, \emph{i.e.}, setting the
compilation flag).
Also observe that one need not accept a tradeoff: calculating
$\log_{10} \eps^{-1} \le 2$ 
and choosing between \acronym{LAMA} and \acronym{LAMA}-size appropriately
would be an easy way to improve performance simultaneously in
ZenoTravel (4 orders of magnitude) and Woodworking ($<2$ orders of magnitude).

Finally, while \acronym{LAMA}-size outperforms \acronym{LAMA}, our
theory of $\eps$-cost traps suggests that cost-based search should
fail even more spectacularly. In an earlier technical report
describing this work \cite{cost-based-arxiv}, we took a much closer look at the travel
domain and present a detailed study of which extensions of
\acronym{LAMA} help it temporarily mask the pernicious effects of
cost-based search. Our conclusion is that both \acronym{LAMA} and
SapaReplan manage to find solutions to problems in the travel domain
despite the use of a cost-based evaluation function by using various
tricks to induce a limited amount of {\em depth-first behavior} in an
A$^*$-framework.  This has the potential effect of delaying
exploration of the $\eps$-cost plateaus slightly, past the discovery
of a solution, but still each planner is ultimately trapped by such
plateaus before being able to find really good solutions.  In other
words, such tricks are mostly serving to mask the problems of
cost-based search (and $\eps$-cost), as they merely delay failure by
just enough that one can imagine that the planner is now effective
(because it returns a solution where before it returned none).  Using
a size-based evaluation function more directly addresses the existence
of cost plateaus, and not surprisingly leads to improvement over the
equivalent cost-based approach --- even with \acronym{LAMA}.

\begin{table}[t]
\caption{\label{tbl:zeno}IPC metric on SapaReplan variants in
  ZenoTravel.}{
\begin{tabular}{|c||c|c|c|c|}
\hline
 & \multicolumn{2}{c|}{2 Cities} & \multicolumn{2}{c|}{3  Cities}\\
\hline
Mode & Score & Rank & Score & Rank\\
\hline
Hybrid                  & 88.8\% & 1& 43.1\% &2\\
Size                    & 83.4\% & 2& 43.7\% &1\\
Size, tie-break on cost & 82.1\% & 3& 43.1\% &2\\
Cost, tie-break on size & 77.8\% & 4& 33.3\% &3\\
Cost                    & 77.8\% & 4& 33.3\% &3\\
\hline
\end{tabular}}

\end{table}

%
%

\mvp
\subsection{Evaluation of Temporal Planning on Temporal Fast Downward}
\label{subsec:tfd}

The problem of $\eps$-cost also exists when attempting minimize
makespan using an makespan-based, ill-behaved evaluation function. In
previous work on this subject Benton~et~al.~(\cite{benton10a})
tested the planner Temporal Fast Downward to see the effect of this;
their results over IPC-2008 domains are shown in
Table~\ref{tbl:makespan}.


\nj{We present some 
results of using useful search operators
as a way to complement surrogate evaluation function. 
 The best-first search in TFD
uses a modified version of the context-enhanced additive
heuristic~\cite{helmert-geffner:icaps-2008} that sums the durations as
costs, $h^{cea}_{dur}$, meaning the heuristic captures a sequential
view of actions. To avoid noise from other search enhancement
techniques, we disabled deferred evaluation for our experiments. To
detect heuristic-useful operators, we used the heuristic of Benton et
al.~\cite{benton10a} that utilizes a Simple Temporal
Network to reschedule heuristic plans. We implemented the useful
operator lookahead discussed earlier into TFD for a planner we call
\tfdu.}

\begin{table*}[t]
  \caption{From useful actions paper.}{
  \centering
  \begin{tabular}{|l||r|r|r|}
  \hline
  Domain &
  cov &
  qual \\
  \hline
    crewplanning-strips & 4 & 4.00 \\
    elevators-numeric & 2 & 2.00 \\
    elevators-strips & 3 & 2.98 \\
    openstacks-adl & 8 & 7.58 \\
    openstacks-strips & 27 & 20.93 \\
    parcprinter-strips & 6 & 5.31 \\
    pegsol-strips & 22 & 21.15 \\
    sokoban-strips & 10 & 10.00 \\
    transport-numeric & 3 & 3.00 \\
    woodworking-numeric & 18 & 16.91 \\
  \hline
    overall & 103 & 93.86 \\
  \hline
  \end{tabular}
  \label{tbl:makespan}
}
\end{table*}

\begin{table*}[t]
  \caption{From g-value plateaus paper.}{
{\small
  \centering
  \begin{tabular}{|l||r|r|r|r||r|r|r|r||r|r|r|r|}
  \hline
  &
  \multicolumn{4}{|c||}{\textbf{$\textbf{f}_{\textbf{c}} = \textbf{g}_{\textbf{c}} + \textbf{h}_{\textbf{c}}$}} &
  \multicolumn{4}{|c||}{\textbf{$\textbf{f}_{\textbf{m}} = \textbf{g}_{\textbf{m}} + \textbf{h}_{\textbf{m}}$}} &
  \multicolumn{4}{|c|}{\textbf{$\textbf{f}_{\textbf{mw}} = \textbf{g}_{\textbf{m}} + \textbf{h}_{\textbf{w}}$}} \\
  Domain &
  $r_g$ &
  $r_f$ &
  cov &
  qual &
  $r_g$ &
  $r_f$ &
  cov &
  qual &
  $r_g$ &
  $r_f$ &
  cov &
  qual \\
  \hline
    crewplanning-strips & 0.03 & 0.55 & 11 & 6.82 & 0.98 & 0.83 & 4 & 4.00 & 0.95 & 0.09 & 12 & 11.99 \\
    elevators-numeric & 0.06 & 0.03 & 4 & 2.41 & 0.57 & 0.27 & 2 & 2.00 & 0.48 & 0.05 & 4 & 3.78 \\
    elevators-strips & 0.07 & 0.05 & 3 & 1.70 & 0.53 & 0.25 & 3 & 2.98 & 0.44 & 0.04 & 4 & 3.92 \\
    openstacks-adl & 0.15 & 0.89 & 30 & 17.80 & 1.00 & 0.88 & 8 & 7.58 & 0.92 & 0.02 & 30 & 28.00 \\
    openstacks-strips & 0.14 & 0.88 & 30 & 17.12 & 0.67 & 0.29 & 27 & 20.93 & 0.71 & 0.06 & 30 & 25.39 \\
    parcprinter-strips & 0.16 & 0.08 & 12 & 5.68 & 0.90 & 0.37 & 6 & 5.31 & 0.76 & 0.09 & 6 & 5.00 \\
    pegsol-strips & 0.17 & 0.09 & 25 & 18.77 & 0.85 & 0.25 & 22 & 21.15 & 0.82 & 0.08 & 26 & 24.04 \\
    sokoban-strips & 0.28 & 0.16 & 11 & 10.18 & 0.78 & 0.32 & 10 & 10.00 & 0.77 & 0.10 & 10 & 9.83 \\
    transport-numeric & 0.23 & 0.06 & 2 & 1.26 & 0.74 & 0.48 & 3 & 3.00 & 0.58 & 0.09 & 3 & 2.78 \\
    woodworking-numeric & 0.08 & 0.12 & 18 & 14.30 & 0.72 & 0.26 & 18 & 16.91 & 0.55 & 0.04 & 19 & 17.64 \\
  \hline
    overall & 0.09 & 0.21 & 146 & 96.04 & 0.84 & 0.50 & 103 & 93.86 & 0.68 & 0.07 & 144 & 132.37 \\
    \hline
  \end{tabular}
  \label{tbl:results}
}}
\end{table*}


We run both TFD and \tfdu\ on the temporal domains from the IPC 2008
(except {\em modeltrain}, as in our tests TFD in general is unable to
solve more than 2 problems in this domain). This benchmark set is
known to contain plateaus on
$g_m$~\cite{benton10a}.\footnote{We used the {\em
    elevators-numeric} and {\em openstacks-adl} variants for our
  results on the respective domains following the IPC-2008 rules by
  using the domain variant that all our planner versions did best in.}
The experiments were run on a 2.7~GHz AMD Opteron processor, with a
timeout of 30 minutes and a
memory~limit~of~2~GB.

  \begin{table}[tbp]
   {\small
    \centering
    \caption{IPC scores when using surrogate useful
  operator in TFD.}{
    \noindent
    {\small 
\begin{tabular}{|l||rr|}
\hline
\textbf{Domain}
& TFD
& \tfdu
\\ \hline
\textbf{Crewplanning}
& 22.44
& 22.57
\\
\textbf{Elevators}
& 13.88
& 15.40
\\
\textbf{Openstacks}
& 25.79
& 27.81
\\
\textbf{Parcprinter}
& 8.73
& 8.47
\\
\textbf{Pegsol}
& 28.73
& 28.81
\\
\textbf{Sokoban}
& 10.93
& 10.79
\\
\textbf{Transport}
& 5.06
& 4.68
\\
\textbf{Woodworking}
& 19.72
& 19.93
\\
\hline
\textbf{Overall}
& \textbf{135.28}
& \textbf{138.48}
\\ \hline
\end{tabular}
}
    \label{table:results-table}
    }
    }
  \end{table}

The IPC scores of our experiment are found in
Table~\ref{table:results-table}.

\includefigure{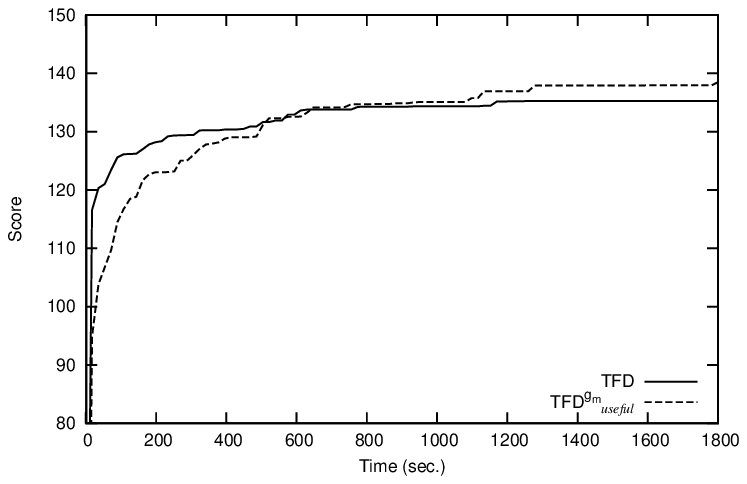}{Anytime behavior showing the change in IPC
  score as time progresses.}

To help see whether the additional computational time spent on
calculating usefulness affects the search, we measured the quality and
coverage across all benchmarks over time using the scoring system from
the International Planning Competition.  The results of this appear in
Figure~\ref{figure:plot-time}. Given the extra time spent on useful
action calculations, it is  unsurprising that the scores for \tfdu\ start
lower than TFD. However, at about 10 minutes \tfdu\ dominates
TFD.

\section{Related Work}

Others have noted issues with using cost-based search and suggested
surrogate approaches. Perhaps most relevant to our own analysis is the
work by Wilt and Ruml~(\cite{wilt2011}). They perform an empirical
analysis similar to ours that shows typical heuristic search suffers as
the result of high ratios between the cost of search operators. They
then prove heuristic error bounds in specific cases of search (i.e.,
using a consistent heuristic and invertible operators) that may be the
beginnings of an explanation of the problem. Further, using our
earlier investigations~\cite{cushing10a,benton10a} as a guide, they
empirically show that search approaches using ``search distance''
(i.e., size) as a component have better performance.

Work done by Thayer and Ruml~(\cite{thayer-ees}) has considered
the issues of blending size and cost in the design of evaluation
functions and search algorithm modifcation. Their work combines a type
of surrogate evaluation over solution size with cost estimates using
different search queues.  
The yet earlier, deeply insightful, work that Thayer, Ruml, and
Westphal build on is that of Pohl into the typical poor behaviour of
A* when given a plethora of equally good choices.  That work led first
to the development of WA* (in a phrase, use $f = g + wh$ as
surrogate), and then on to depth-dependent weighting variants of A*
(let $w$ vary somewhat intelligently with depth, i.e., penalize
\emph{size})~\cite{pohl70,pohl73}.  To quote Pohl quoting
Kowalski~\cite{kowalski72}: ``...have a critical defect.  They
[$\text{A}^*$] investigate {\em all} equally meritorious alternative
paths.  \emph{In difficult problem spaces where any solution path is
  desired, this procedure is inappropriately breadth first.}''
(emphasis added).


Dechter and Pearl~(\cite{dechter85}) give a highly technical
account of the properties of generalized best-first search strategies,
focusing on issues of computational optimality, but mostly from the
perspective of search constrained to proving optimality in the path
metric.  Additionally, Richter and Westphal~(\cite{lama-journal})
have a thorough empirical analysis of cost issues in standard planning
benchmarks in their planner LAMA. Subsequently LAMA was modified to
include a surrogate, size-based greedy best-first search search to
find a first solution before beginning the same search approach as its
previous version~\cite{lama2011}.






\mvp
\section{Conclusion}

Several researchers have noted the troublesome behavior of
  cost-based search in many planning benchmarks, suggesting several
  {\em ad hoc} work arounds for it. 
In this paper, we tried to provide a general explanation for the
malady.  Specifically,  we argued that
the origins of this problem can be traced back to the fact that most
planners that try to optimize cost also use cost-based evaluation
functions (i.e., $f(n)$ is a cost estimate). We showed that cost-based
evaluation functions become ill-behaved whenever there is a  wide
variance in action costs; something that is all too common  in
planning domains.  The general solution to this malady is what we call
a {\em surrogate search}, where a surrogate evaluation function that
does not directly track the cost objective, and is resistant to
cost-variance, is used. To shed light on the art of devising good
surrogate evaluation functions, we proposed cost-surrogates in
cost-based planning and makespan-surrogates in temporal planning. 
For the former, we  discussed some compelling choices for
surrogate evaluation functions that are based on size rather than
cost, and showed that they are immune to the difficulties faced by
cost-based search.   Of particular practical interest is a cost-sensitive version of
size-based evaluation function — where the heuristic estimates the
size of cheap paths, as it provides attractive quality vs. speed
tradeoffs. For temporal planning, we proposed a method that
significantly improves upon the surrogate evaluation function by tracking
operator usefulness. 
Our empirical evaluations with  state-of-the-art planners
demonstrate the effectiveness of surrogate search.
\nrao{A worthwhile direction for future work is finding other 
surrogate evaluation functions that strike an even  better
  balance between resistance to $\eps$-cost traps, and focus on the objective.}

\bibliographystyle{plain}
\bibliography{fsize}
\end{document}